\documentclass[10pt,twocolumn,letterpaper]{article}
\usepackage{cvpr}              

\usepackage{graphicx}
\usepackage{amsmath}
\usepackage{amssymb}
\usepackage{dsfont}
\usepackage{booktabs}
\usepackage{algorithm}
\usepackage{algpseudocode}
\usepackage{algcompatible}

\usepackage{multirow}
\usepackage{booktabs}
\usepackage{makecell}
\usepackage{xcolor}

\usepackage[pagebackref,breaklinks,colorlinks]{hyperref}
\usepackage[accsupp]{axessibility}

\usepackage[capitalize]{cleveref}
\crefname{section}{Sec.}{Secs.}
\Crefname{section}{Section}{Sections}
\Crefname{table}{Table}{Tables}
\crefname{table}{Tab.}{Tabs.}

\usepackage[normalem]{ulem}

\def\cvprPaperID{5593} 
\def\confName{CVPR}
\def\confYear{2022}

\begin{document}
\newcommand\ChangeRT[1]{\noalign{\hrule height #1}}

\title{Perturbed and Strict Mean Teachers for Semi-supervised Semantic Segmentation}
\author{
\parbox{0.7\linewidth}{\centering $\quad$ Yuyuan Liu\textsuperscript{\rm 1} $\quad$ Yu Tian\textsuperscript{\rm 1} $\quad$    Yuanhong Chen\textsuperscript{\rm 1} $\quad$ Fengbei Liu\textsuperscript{\rm 1}$\newline$  $\quad$  Vasileios Belagiannis\textsuperscript{\rm 2} $\quad$ $\quad$ $\quad$ Gustavo Carneiro\textsuperscript{\rm 1} \\   
\textsuperscript{\rm 1} Australian Institute for Machine Learning, University of Adelaide \\
\textsuperscript{\rm 2} Universit\"at Ulm, Germany} 
}
\maketitle
\begin{abstract}

Consistency learning using input image, feature, or network perturbations has shown remarkable results in semi-supervised semantic segmentation, but this approach can be seriously affected by inaccurate predictions of unlabelled training images. There are two consequences of these inaccurate predictions: 1) the training based on the ``strict'' cross-entropy (CE) loss can easily overfit prediction mistakes, leading to confirmation bias; and 2) the perturbations applied to these inaccurate predictions will use potentially erroneous predictions as training signals, degrading consistency learning. In this paper, we address the prediction accuracy problem of consistency learning methods with novel extensions of the mean-teacher (MT) model, which include a new auxiliary teacher, and the replacement of MT's mean square error (MSE) by a stricter confidence-weighted cross-entropy (Conf-CE) loss. The accurate prediction by this model allows us to use a challenging combination of network, input data and feature perturbations to improve the consistency learning generalisation, where the feature perturbations consist of a new adversarial perturbation. Results on public benchmarks show that our approach achieves remarkable improvements over the previous SOTA methods in the field.\footnote{Supported by Australian Research Council through grants DP180103232 and FT190100525.}  Our code is available at \url{https://github.com/yyliu01/PS-MT}. \vspace{-15pt}  

\end{abstract}

\section{Introduction}
\label{sec:intro}

Semantic segmentation is an essential pixel-wise classification task that has reached remarkable success in recent years. However, the training of such a task is known to be data-hungry, where the labelling process is particularly costly and time-consuming~\cite{ouali2020semi}. To tackle this limitation, semi-supervised semantic segmentation has become an important research direction that has drawn a growing attention recently~\cite{ouali2020semi, chen2021semi, ke2020guided}. 
This problem relies on a small set of pixel-level labelled images and a large set of unlabelled images, where both types of images are drawn from the same data distribution. 
The challenge is how to extract additional and useful training signal from the unlabelled images to allow the training of the model to generalise beyond the small labelled set.

\begin{figure}[t] 
    \centering
    \begin{subfigure}[b]{0.47\linewidth}
         \centering    
            \includegraphics[width=1.\linewidth]{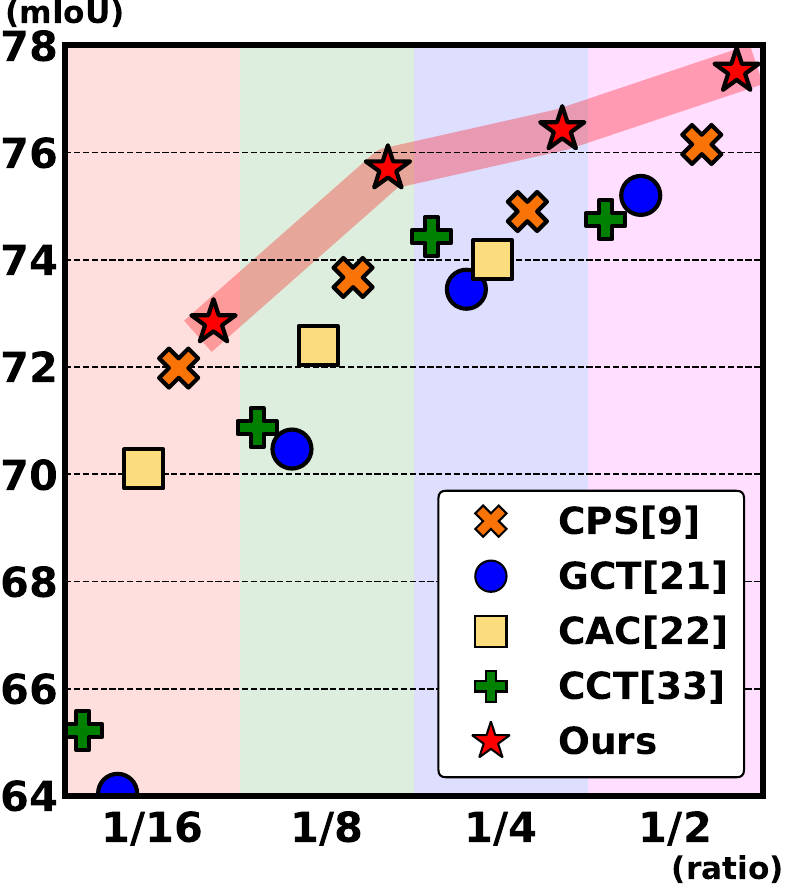}
    \end{subfigure}
        \begin{subfigure}[b]{0.49\linewidth}
         \centering     \includegraphics[width=1.\linewidth]{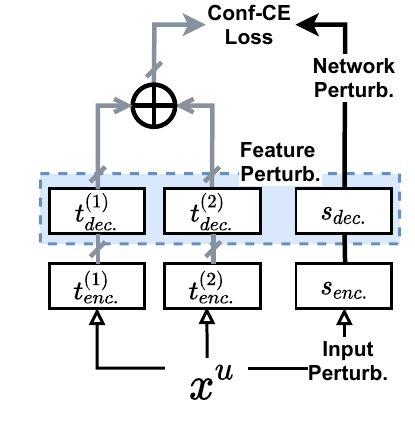}
    \end{subfigure}
    \caption{In (a), we compare our method (red star) to the current SOTA on \textit{Pascal VOC 2012} under different partition protocols based on the augmented set~\cite{hariharan2011semantic}, and (b) shows a simple diagram of our consistency-based model, which includes two mean teachers (denoted by the encoders $t^{(1)}_{enc.}$ and $t^{(2)}_{enc.}$ and decoders $t^{(1)}_{dec.}$ and $t^{(2)}_{dec.}$) that produce accurate segmentation maps for unlabelled images $x^u$ and the student (encoder $s_{enc.}$ and decoder $s_{dec.}$), with three types of perturbations (network, feature and input image) that are used in the minimisation of our strict Conf-CE loss.\vspace{-11pt}}
    \label{fig: brief_talk}
\end{figure}

Current state-of-the-art (SOTA) semi-supervised semantic segmentation models are based on consistency learning, 
which enforces the agreement between the outputs from different views of unlabelled images~\cite{ouali2020semi, zou2020pseudoseg, french2019semi,chen2021semi}. 
These different views can be obtained via perturbations applied to the input image with data augmentation~\cite{zou2020pseudoseg} or to the feature space with noise injection~\cite{ouali2020semi}. Another way of obtaining different views is with network perturbation, which encourages similar predictions between multiple models trained from different initialization, and has been shown to enable better consistency regularization than input image and feature perturbations~\cite{chen2021semi, ke2020guided}.
One potential weakness of consistency learning is that it assumes accurate predictions for unlabelled images, such that the perturbation does not push the image feature to the wrong side of the true (hidden) classification decision boundary. 
Unfortunately, in practice this assumption is not always met by SOTA methods, making the training signal of consistency learning methods potentially incorrect. 
This problem is exacerbated for consistency learning based on network perturbation because incorrect predictions from one model will deteriorate the training for the other model, and vice versa.
Another consequence of these inaccurate predictions is that consistency learning methods that rely on a ``strict'' cross-entropy (CE) loss can easily overfit prediction mistakes, which can lead to confirmation bias.


In this paper, we address the prediction accuracy problem of consistency based methods by extending the mean teacher (MT) model~\cite{tarvainen2017mean,french2019semi,chen2021semi, ke2020guided} with a new auxiliary teacher, and the replacement of MT's means square error (MSE) loss by a stricter confidence-weighted CE loss (Conf-CE) that has better training convergence.
These accurate predictions  enable the use of more challenging perturbations, combining input image, feature and network perturbations to improve the generalisation of consistency learning.
Furthermore, we propose a new type of adversarial feature perturbation that learns the perturbation to be applied to the student model using virtual adversarial training~\cite{miyato2018virtual} from the teachers (T-VAT), instead of injecting different types of noise in the image features~\cite{ouali2020semi}.
To summarise, our contributions are:
\begin{itemize}
\item New consistency based semi-supervised semantic segmentation MT model designed to improve the segmentation accuracy of unlabelled training images with a new auxiliary teacher and a replacement of MT's MSE loss by a stricter confidence-weighted CE loss (Conf-CE) that allows stronger convergence and overall better training accuracy;
\item A new challenging combination of input data, feature and network perturbations to improve model generalisation; and
\item A new type of feature perturbation, called T-VAT, based on an adversarial noise learned from the both teachers of our MT model and applied to the student model, which results in the generation of challenging noise to promote an effective training of the student model.
\end{itemize}
Our experimental evaluation shows that our approach achieves the best results on \textit{Pascal VOC 2012}~\cite{everingham2015pascal}. 
Our approach also shows the best performance on Cityscapes~\cite{cordts2016Cityscapes}. 


\section{Related Work}

Below, we first discuss supervised semantic segmentation, then semi-supervised learning, and then we describe pseudo-labelling and consistency-based SSL methods.

\noindent \textbf{Supervised semantic segmentation} consists of classifying all image pixels into a closed set of visual classes. 
Current models are based on fully convolutional neural networks (FCN)~\cite{long2015fully,badrinarayanan2017segnet, minaee2021image} and extensions that explore: 1) multi-scale  aspects of the image\cite{dai2015convolutional, lin2016efficient}, 
2) pyramidal feature maps~\cite{zhao2017pyramid, ghiasi2016laplacian, he2019adaptive}, 3) dilated convolutions~\cite{yu2015multi, chen2018encoder, chen2017rethinking}, and 4) attention mechanisms~\cite{chen2016attention, li2019expectation}.  
SOTA semi-supervised semantic segmentation models rely on the supervised semantic segmentation models DeeplabV3+~\cite{chen2017deeplab} and 
PSPNet~\cite{zhao2017pyramid} as backbone architectures.

\noindent  \textbf{Semi-supervised learning (SSL)} trains a model using labelled and unlabelled images.
Current SSL solutions are formulated based on three assumptions~\cite{van2020survey}: 1) smoothness: similar images have similar labels; 2) low-density: decision boundary does not pass through high-density areas of the feature space; and 3) manifold: samples on the same low-dimensional manifold embedded in the feature space have the same label.
SSL methods can be loosely classified  into \textbf{pseudo-label based SSL}~\cite{berthelot2019mixmatch, sohn2020fixmatch, 9207304} and \textbf{consistency based SSL}~\cite{laine2016temporal,tarvainen2017mean, polyak1992acceleration}, with 
the former generally presenting worse accuracy than the latter. We believe that this is due to the fact that pseudo-label methods disregard part of the unlabelled training set during training, which can reduce their generalisation ability.
Below, we focus on consistency based SSL given its superior accuracy on public benchmarks.

\begin{figure*}[t!]
    \centering
    \hspace*{1.2cm}\includegraphics[width=.86\textwidth]{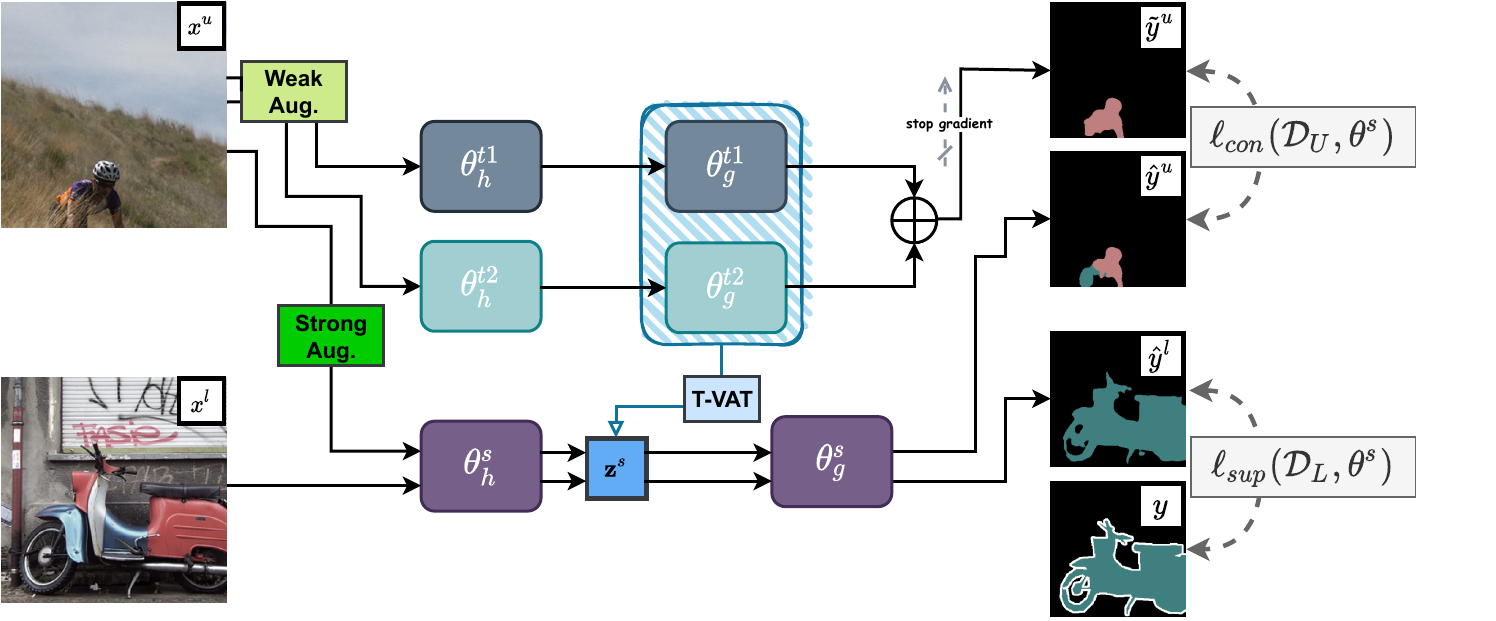}
    \caption{Illustration of our approach. The unlabelled image $\mathbf{x}^{u}$ is weakly augmented for the mean teachers (encoders parameterised by $\theta_h^{t1,t2}$ and decoders by $\theta_g^{t1,t2}$) that jointly predict the segmentation $\tilde{\mathbf{y}}^{u}$.
    The same unlabelled image 
    is strongly augmented for the student (with encoder $\theta_h^{s}$ and decoder $\theta_g^{s}$) that suffers  T-VAT feature perturbation before predicting the segmentation $\hat{\mathbf{y}}^{u}$.
    This prediction of the unlabelled image and the prediction $\hat{\mathbf{y}}^{l}$ of the labelled image $\mathbf{x}^{l}$ (also perturbed by T-VAT) are used to minimise the consistency loss $\ell_{con}(.)$ and the supervised loss $\ell_{sup}(.)$, respectively, to train the student. The mean teachers are trained with EMA of the student model. 
    }
\vspace{-10pt}
\end{figure*}

\noindent \textbf{Consistency-based SSL methods} aims to enforce the agreement between the predictions of perturbed unlabelled images, where perturbations can be applied to the input image, to the feature representation, or to the network. 
The effectiveness of consistency-based SSL depends on the prediction accuracy for the unlabelled images and the perturbations to challenge the model training. 
In general, more challenging perturbations target an improved generalisation, but if this perturbation is applied to inaccurate predictions, it can cause the consistency-based method to learn from wrong labels.  
Prediction accuracy can be improved in many ways, but a simple model ensemble strategies~\cite{laine2016temporal,tarvainen2017mean}.
Perturbations can be applied to the input image~\cite{zou2020pseudoseg}, feature representation~\cite{ouali2020semi} or network~\cite{chen2021semi, ke2020guided}.
Independently of how they are applied, perturbations tend to be more effective when they challenge the classification process by, for example, moving perturbed feature closer to true (but hidden) classification boundaries,
such as with virtual adversarial training (VAT)~\cite{miyato2018virtual}. 

\noindent \textbf{Consistency-based semi-supervised semantic segmentation methods} have shown more competitive results~\cite{french2019semi, ouali2020semi, chen2021semi} than pseudo-label approaches. 
Among the SOTA consistency-based semi-supervised semantic segmentation methods, PseudoSeg~\cite{zou2020pseudoseg} 
relies on a new pseudo-labelling strategy and data augmentation consistency training to calibrate pseudo-labels, but its dependence on the generally inaccurate class activation maps can lead to poor training performance. 
Cross-consistency training (CCT)~\cite{ouali2020semi} applies different types of feature perturbations to enforce the agreement between their semantic segmentation results and the segmentation from the non-perturbed feature. Although the feature perturbations used in CCT are effective, more targeted and accurate adversarial noise can be more helpful for the consistency regularization.
Other methods have explored network perturbation~\cite{feng2020semi, ke2020guided, mendel2020semi, chen2021semi}, where consistency is enforced between the responses of differently initialised models.
Perturbation models depend on the ability of the models to produce accurate segmentation results, and as mentioned above, such ability can be improved with the use of model ensembling~\cite{tarvainen2017mean}. 
French et al.~\cite{french2019semi} 
explore model ensembling~\cite{tarvainen2017mean} together with network perturbation and input image perturbation~\cite{yun2019cutmix}. This is one of the closest methods to our proposal, but we add a more effective model ensembling with multiple mean teachers, and a new adversarial feature perturbation with VAT~\cite{miyato2018virtual} and challenging input image perturbation with CutMix~\cite{yun2019cutmix} and Zoom In/Out~\cite{lin2018multi, chen2016attention} 
data augmentation. 
Also comparing with~\cite{french2019semi}, the more accurate segmentation results produced by our multiple mean teachers allows us to train the model for unlabelled images with the CE loss instead of the MSE used in~\cite{french2019semi}, providing better training convergence and accuracy. 



\section{Method} 
\label{sec:method}

Before we describe our model and training process, we introduce our dataset for semi-supervised semantic segmentation. 
We have a small labelled training set $\mathcal{D}_L = \{ (\mathbf{x}_i,\mathbf{y}_i)\}_{i=1}^{|\mathcal{D}_L|}$, where $\mathbf{x}_i \in \mathcal{X} \subset \mathbb{R}^{H \times W \times C}$ is the input image of size $H \times W$ with $C$ colour channels, and $\mathbf{y}_i  \in \mathcal{Y} \subset \{0,1\}^{H \times W \times Y}$ is the segmentation map, with the number of visual classes denoted by $Y$.  
We also have a large unlabelled training set $\mathcal{D}_U = \{ \mathbf{x}_i\}_{i=1}^{|\mathcal{D}_U|}$ is provided, with $|\mathcal{D}_L| << |\mathcal{D}_U|$. 
These datasets are used to train our proposed MT model with an auxiliary teacher, described in Sec.~\ref{sec:multiple_mean_teacher_student_model}.
The training of our new MT model, exploring network, feature and input image perturbations, and a strict Conf-CE loss is described in Sec.~\ref{sec:consistency_loss}. 
\subsection{Multiple Mean Teachers and Student Models}
\label{sec:multiple_mean_teacher_student_model}

As explained in Sec.~\ref{sec:intro}, we aim to improve the accuracy of the segmentation of unlabelled training images. To achieve that, we propose the inclusion of an auxiliary teacher, exploring the idea of a double ensembling process to improve segmentation accuracy, namely the 
ensemble of teacher models, each representing a temporal ensemble of the student model~\cite{tarvainen2017mean}. 
The teachers and student models have the same network structure denoted by $f_\theta:\mathcal{X} \to \mathbb{R}^{H \times W \times Y}$, where $\theta$ is the model parameter.
This model is decomposed into an encoder $h_{\theta_h}:\mathcal{X} \to \mathcal{Z}$ and a decoder $g_{\theta_g}:\mathcal{Z} \to \mathcal{Y}$, where $\mathcal{Z} \subset \mathbb{R}^Z$ represents the feature space of dimension $Z$. Hence, $f_{\theta} = g_{\theta_g} \circ h_{\theta_h}$, where $\theta = \{ \theta_g,\theta_h \}$. The probability output of the network is achieved by applying the pixel-wise softmax function $\sigma(.)$ over the $Y$ classes, as in
$p_{\theta}(\mathbf{x}) = \sigma(f_{\theta}(\mathbf{x}))$.
The multiple mean teacher-student model is represented with the respective parameter superscripts: $\theta^{t1} = \{ \theta^{t1}_g,\theta^{t1}_h \}$ and $\theta^{t2} = \{ \theta^{t2}_g,\theta^{t2}_h \}$ for the teachers, and $\theta^{s} = \{ \theta^{s}_g,\theta^{s}_h \}$ for the student.

\subsection{Training with Multiple Perturbations and a Strict Confidence-weighted CE Loss}
\label{sec:consistency_loss}

In this section, we present the training process of our new MT model using a confidence-weighted CE loss, which is optimised with perturbations to 
the network, feature representations and input images.

\textbf{Training.} The full training loss for the student model is
\begin{equation}
    \ell(\mathcal{D}_L,\mathcal{D}_U,\theta^s) =  \ell_{sup}(\mathcal{D}_L,\theta^s) + 
    \beta \ell_{con}(\mathcal{D}_U,\theta^s),
\label{eq:main_loss}
\end{equation}
where the first loss is the supervised segmentation loss, defined as:
\begin{equation}
\resizebox{.95\hsize}{!}{$
\ell_{sup}(\mathcal{D}_L,\theta^s) = \frac{1}{|\mathcal{D}_L||\Omega|}\sum_{(\mathbf{x},\mathbf{y})\in\mathcal{D}_L} 
\sum_{\omega \in \Omega} \ell(\mathbf{y}(\omega),p_{\theta^{s}}(\mathbf{x})(\omega)),%
$}
    \label{eq:loss_sup}
\end{equation}
where $\Omega$ is the image lattice of size $H \times W$, and $\ell(.)$ denotes the CE loss between the annotation $\mathbf{y}$ and
segmentation prediction from $p_{\theta^{s}}(\mathbf{x})$ at pixel address $\omega \in \Omega$. 
The second term in~\eqref{eq:main_loss} is the consistency loss, denoted by the confidence weighted CE loss (\textbf{Conf-CE}), with
$\beta \in \mathbb{R}$ weighting its contribution and its definition being as follows:
\begin{equation} \label{eq:loss_semi}
\resizebox{.95\hsize}{!}{$%
\ell_{con}(\mathcal{D}_U,\theta^s) = \frac{1}{|\mathcal{D}_U||\Omega|}\sum_{\mathbf{x}\in\mathcal{D}_U}
\sum_{\omega \in \Omega} c(\omega)\ell(\tilde{\mathbf{y}}(\omega),p_{\theta^{s}}(\mathbf{x})(\omega)),
$}
\end{equation}
where $\ell(.)$ represents the CE loss,
$\omega$ denotes the pixel address of the output lattice $\Omega$ of the segmentation map,
$\tilde{\mathbf{y}}(\omega) \in \{0,1\}^{Y}$ is the segmentation prediction from the teacher models at $\omega$, 
$p_{\theta^{s}}(\mathbf{x})(\omega) \in [0,1]^{Y}$ is the student model segmentation prediction at $\omega$, 
and
$c({\omega})\in[0,1]$ represents the segmentation prediction confidence from the teacher models at $\omega$, defined below in~\eqref{eq:teacher_ensemble_result}. 


The \textbf{network perturbation} 
is obtained from the  predictions of the mean teacher models and the student model. 
The soft segmentation map produced by the ensemble of the mean teachers is estimated as:
\begin{equation}
    \hat{\mathbf{y}} = \sigma (0.5 \times( f_{\theta^{t1}}(\mathbf{x}) + f_{\theta^{t2}}(\mathbf{x}))),
    \label{eq:teacher_ensemble_result}
\end{equation}
where $\sigma(.)$ denotes the softmax function. The hard segmentation prediction by the ensemble of teachers, denoted by $\tilde{\mathbf{y}} \in \mathcal{Y}$ in~\eqref{eq:loss_semi}, is obtained from the one-hot representation computed from $\hat{\mathbf{y}} \in [0,1]^{H \times W \times Y}$ in~\eqref{eq:teacher_ensemble_result}.
The segmentation prediction confidence $c(\omega)$ in~\eqref{eq:loss_semi} is computed with $c(\omega)=\tilde{\mathbf{y}}(\omega)^{\top}\hat{\mathbf{y}}(\omega) \times \mathbb{I}(\tilde{\mathbf{y}}(\omega)^{\top}\hat{\mathbf{y}}(\omega) > \tau)$, where $\mathbb{I}(.)$ denotes an indicator function and $\tau$ is a minimum confidence to enable a value larger than zero for $c(\omega)$.
Following the MT framework~\cite{tarvainen2017mean}, while the student model is trained via stochastic gradient descent (SGD) to minimise the cost function in~\eqref{eq:loss_semi}, both teacher models are trained with exponential moving average (EMA)~\cite{tarvainen2017mean} of the student model and batch norm (BN) parameters~\cite{cai2021exponential}, with:
\begin{equation} \label{eq:ema}
    \theta^{k} = \gamma \times \theta^{k} + (1-\gamma) \times \theta^s,
\end{equation}
where $k\in \{t_1, t_2\}$, and $\gamma \in (0, 1)$ controls the transfer weight between epochs.  
For the training of teacher models, we update the parameters of only one of the two teachers at each training epoch. 


The \textbf{feature perturbation} 
consists of a challenging adversarial feature perturbation that is designed to violate the cluster, or low-density, assumption~\cite{ouali2020semi,van2020survey} by pushing the image features, computed from the model encoder, toward the classification boundaries in the feature space. 
One effective way to produce such adversarial feature noise is with virtual adversarial training (VAT)~\cite{miyato2018virtual} that optimises a perturbation vector to maximise the divergence between correct and adversarial classification.
Current methods  
estimate the adversarial noise using 
the same single network where the consistency loss will be applied~\cite{ouali2020semi}.
However, the use of VAT to perturb the training of MT in semi-supervised semantic segmentation is new, to the best of our knowledge.
In an MT model, the feature perturbation can be applied to the student model,
but given that it has less accurate predictions than the teacher model, this approach may not be conducive to effective training. 
Hence we propose to estimate the adversarial noise using the more accurate teachers, and then apply this estimated noise to the feature of the student model -- we call this feature perturbation \textbf{T-VAT}.
The student output to be used in the loss in~\eqref{eq:loss_semi} is $p_{\theta^s}(\mathbf{x}) = \sigma(g_{\theta_g^s}(h_{\theta_h^s}(\mathbf{x})+\mathbf{r}_{adv}))$, where 
the adversarial feature perturbation $\mathbf{r}_{adv} \in \mathcal{Z}$ is estimated from the response of the ensemble of teacher models with:
\begin{equation}
\resizebox{.95\hsize}{!}{$
\begin{aligned}
        \text{maximise} \;\; & d \Big ( \sigma\big (0.5 \times  (g_{\theta_g^{t1}}(\mathbf{z}^{s})+ g_{\theta_g^{t2}}(\mathbf{z}^{s})) \big ), \\
        & \;\;\;\;  \sigma \big(0.5\times (g_{\theta_g^{t1}}(\mathbf{z}^{s} + \mathbf{r}_{adv}) + g_{\theta_g^{t2}}(\mathbf{z}^{s} + \mathbf{r}_{adv}) \big ) \Big ), \\
        \text{subject to} \;\; & ||\mathbf{r}_{adv}||_2 <= \epsilon,
\end{aligned}%
$}
\label{eq:teacher_ensemble_adversarial_result}
\end{equation}
where $\mathbf{z}^{s}=f_{\theta^{s}}(\mathbf{x})$,
$d(.)$ is the the sum of the pixel-wise Kullback-Leibler (KL) divergence between the original and perturbed pixel predictions.

The \textbf{input image perturbation} is based on the weak-strong augmentation pairs~\cite{lee2013pseudo},  where weak augmentations (image flipping, cropping and scaling) are applied to the images  to be processed by the teacher models, and in addition to those weak augmentation, strong augmentations~\cite{chen2021semi, ke2020guided} (e.g., colour jitter, randomise grayscale and blur) are applied to the images fed to the student model to improve the overall generalization capability. 

On top of the strong augmentations, we also apply the CutMix~\cite{yun2019cutmix} and Zoom In/Out~\cite{lin2018multi, chen2016attention} data augmentations to the student model images.
As defined in~\cite{french2019semi}, the CutMix augmentation is achieved  by applying a binary mask $\mathbf{m} \in \{0,1\}^{H \times W}$ that combines two images using the function $\mu(\mathbf{x}_i,\mathbf{x}_j,\mathbf{m})=(1-\mathbf{m})\odot\mathbf{x}_i + \mathbf{m} \odot \mathbf{x}_j$. We can apply CutMix by combining two input images and minimise the consistency loss~\eqref{eq:loss_semi} with the prediction from~\eqref{eq:teacher_ensemble_result}~\cite{chen2021semi} (referred to as \textit{CutMix before prediction}), or we can minimise the consistency loss using the CutMix combination of the images and their predictions, as in
\begin{equation}
    \ell(\mu(\tilde{\mathbf{y}}_i,\tilde{\mathbf{y}}_j,\mathbf{m}),p_{\theta^s}(\mu(\mathbf{x}_i,\mathbf{x}_j,\mathbf{m}))),
    \label{eq:cutmix_loss}
\end{equation}
with $\tilde{\mathbf{y}}$ defined in~\eqref{eq:teacher_ensemble_result}.
The perturbation used in~\eqref{eq:cutmix_loss} is referred to as \textit{CutMix after prediction}, which we argue to produce a cleaner prediction for the consistency loss than \textit{CutMix before prediction} because its prediction does not contain the artifacts introduced by the prediction from the CutMix images. 
The Zoom In/Out augmentation~\cite{lin2018multi, chen2016attention} is defined by the function $\zeta(\mathbf{x},s )$ that zooms in or out the image using the parameter $s  \in \mathbb{R}_+$. The input image consistency loss in~\eqref{eq:loss_semi} for the zoom in/out augmentation for the ensemble results of teacher models is defined by
\begin{equation}
    \ell(\zeta(\tilde{y}, s),p_{\theta^s}(\zeta(\mathbf{x},s ))),
\end{equation}
and $\tilde{y}$ is defined in~\eqref{eq:teacher_ensemble_result}.

The segmentation loss $\ell(.)$ in~\eqref{eq:loss_semi} for previous consistency-based semi-supervised semantic segmentation models~\cite{french2019semi, chen2021semi}
is usually based on the L2 loss. 
Even though L2 loss is known to be robust, which is advantageous when dealing with the noisy predictions produced by consistency-based methods, it is also known to have poor converge and to possibly lead to vanishing gradients. 
Given the reliability of the segmentation predictions produced by our extended MT model, we instead use the more effective cross entropy (CE) loss, constrained to be computed at regions of high-confidence segmentation results, represented by $c(\omega)$ in~\eqref{eq:loss_semi}, following the strategy applied in self-training approaches~\cite{yuan2021simple, he2021re, yang2021st++}.

\textbf{Inference.} The semantic segmentation of a test image is obtained from the teachers, as described in~\eqref{eq:teacher_ensemble_result}. 

\section{Experiments}
We firstly introduce the experimental setting that we used to evaluate our approach. In Sec.~\ref{sec:4.2} we demonstrate our approach for both datasets under different partition protocols by comparing them with the supervised baselines and other previous SOTA approaches. Moreover, we also carry out detailed results based on the few supervision studies in Sec.~\ref{sec:4.3}. Lastly, we perform the ablation study in Sec.~\ref{sec:4.4} and demonstrate an extension experiment based on the exploring of the image-level data in Sec.~\ref{sec:pseudo_label_consistency}.
\subsection{Experimental Setup} \label{sec: set-up}
\label{sec:4.1}
\begin{table*}[h!]
\caption{\textbf{Comparison with SOTA  on Pascal VOC 2012.} All  approaches are based on the DeeplabV3+ architecture, under different partition protocols~\cite{chen2021semi}. The $*$ indicates the approaches re-implemented by~\cite{chen2021semi}. Best results are in bold.} \label{table:partition-voc}
\centering
\resizebox{0.93\textwidth}{!}{\begin{tabular}{!{\vrule width 1pt}c!{\vrule width 1pt}c|c|c|c!{\vrule width 1pt}c|c|c|c!{\vrule width 1pt}} 
\specialrule{1pt}{0pt}{0pt}
\multirow{2}{*}{Method}      & \multicolumn{4}{c!{\vrule width 1pt}}{ResNet-50~}               & \multicolumn{4}{c!{\vrule width 1pt}}{ResNet-101}                \\ 
\cline{2-9}
                             & 1/16(662) & 1/8(1323) & 1/4(2646) & 1/2(5291) & 1/16(662) & 1/8(1323) & 1/4(2646) & 1/2(5291)  \\ 
\specialrule{1pt}{0pt}{0pt}
MT*~\cite{tarvainen2017mean} & 66.77     & 70.78    & 73.22     & 75.41     & 70.59     & 73.20      & 76.62      & 77.61      \\
French et al.*~\cite{french2019semi}  
& 68.90     & 70.70     & 72.46     & 74.49     & 72.56     & 72.69     & 74.25     & 75.89      \\
CCT*~\cite{ouali2020semi} 
& 65.22     & 70.87     & 73.43     & 74.75     & 67.94     & 73.00     & 76.17     & 77.56      \\

GCT*~\cite{ke2020guided} 
& 64.05     & 70.47     & 73.45     & 75.20     & 69.77     & 73.30     & 75.25     & 77.14      \\

ECS~\cite{mendel2020semi}  
& -     & 67.38     & 70.70     & 72.89          & -     & -     & -     & -      \\
CPS~\cite{chen2021semi} 
& 71.98     & 73.67     & 74.90     & 76.15     & 74.48     & 76.44     & 77.68     & 78.64      \\
CAC~\cite{lai2021semi} 
& 70.10      & 72.40      & 74.00      & -         & 72.40      & 74.60      & 76.30      & -          \\

\specialrule{1pt}{0pt}{0pt}
Ours                         &     \textbf{72.83}     & \textbf{75.70}      &     \textbf{76.43}      &    \textbf{77.88}       &     \textbf{75.50}      & \textbf{78.20}      &  \textbf{78.72}         &   \textbf{79.76}         \\
\specialrule{1pt}{0pt}{0pt}
\end{tabular}}

\end{table*}

\begin{table}
\centering
\caption{\textbf{Comparison with SOTA on Cityscapes} under different partition protocols. All the approaches are based on DeeplabV3+. The $\dagger$ indicates the experimental settings (e.g., supervised loss, resolution) follow CPS~\cite{chen2021semi}.} 
\label{table:partition-city}
\resizebox{0.92\linewidth}{!}{
\begin{tabular}{!{\vrule width 1pt}l!{\vrule width 1pt}c|c|c|c!{\vrule width 1pt}} 
\specialrule{1pt}{0pt}{0pt}
Method                               & Backbone  & 1/8   & 1/4   & 1/2    \\ 
\specialrule{1pt}{0pt}{0pt}
ECS~\cite{mendel2020semi}                                  & ResNet50  & 67.38 & 70.70 & 72.89  \\
CAC~\cite{lai2021semi}                                  & ResNet50  & 69.70  & 72.70  & -      \\
Ours                                 & ResNet50  & 74.37 & 75.15 &  76.02      \\ 
\specialrule{1pt}{0pt}{0pt}
\multirow{2}{*}{Ours (sliding eval.)} & ResNet50  & 75.76 & 76.92 &  77.64      \\
                                     & ResNet101 & 76.89 & 77.60 & 79.09  \\
\specialrule{1pt}{0pt}{0pt}
GCT~\cite{ke2020guided}$^\dagger$     & ResNet50  & 71.33 & 75.30 & 77.09      \\
CPS~\cite{chen2021semi}$^\dagger$     & ResNet50  & 76.61 & 77.83 & 78.77      \\
Ours$^\dagger$             & ResNet50  & 77.12 & 78.38 &  79.22      \\ 
\specialrule{1pt}{0pt}{0pt}
\end{tabular}}
\vspace{-10pt}
\end{table}
\textbf{Datasets.} \textit{Pascal VOC 2012}~\cite{everingham2015pascal} is the standard dataset used for evaluating the performance of the semi-supervised segmentation models. The dataset contains more than $13,000$ images with $21$ classes, providing $1,464$ images with pixel-level labels for training, $1,449$ images for validation and $1,456$ for testing. Following previous papers~\cite{ouali2020semi, chen2021semi}, we adopt the additional labels from~\cite{hariharan2011semantic} and our entire training set contains $10,582$ images. 
Note that the labels from~\cite{hariharan2011semantic} are of low-quality, and may contain noise.
\textit{Cityscapes}~\cite{cordts2016Cityscapes} is the urban driving scene dataset, which consists of $2,975$ images for training, $500$ for validation and $1,525$ for testing. Each image in the dataset has resolution $2,048 \times 1,024$, and there are $19$ classes in total. 

Following \cite{ke2020guided, chen2021semi}, we random split the full labelled set in \textit{Pascal VOC 2012} and \textit{Cityscapes} with different ratios. We also provide the results based on the official labelled set (with $1,464$ images for \textit{Pascal VOC 2012})~\cite{zou2020pseudoseg, yuan2021simple}.

\textbf{Implementation details.}
Most results are based on using our method to train the DeeplabV3+~\cite{chen2017deeplab} model. We load the ImageNet pre-trained checkpoint, and the segmentation heads are initialized randomly. Following previous papers~\cite{ouali2020semi, he2021re, chen2021semi}, we utilise the following polynomial learning-rate decay: $(1-\frac{\text{iter}}{\text{max\_iter}})^{0.9}$. We also test our method on PSPNet~\cite{ouali2020semi, lai2021semi} to show the generalization of our approach.

During training, we apply data augmentation with random scaling in $\{0.5, 0.75, 1.25\}$ and random flipping of both labelled and unlabelled images. 
On \textit{Pascal VOC 2012}, we crop images to $512 \times 512$ pixels for DeeplabV3+, train for $80$ epochs with base learning rate set to $0.01$, using batch size $32$,
for both labelled and unlabelled images, following~\cite{chen2021semi}.
For PSPNet, we follow~\cite{he2021re} and crop images to $321 \times 321$ pixels and use batch size $8$. 
On \textit{Cityscapes}, due to hardware limitation, we crop images to $712 \times 712$ pixels, train for $450$ epochs with base learning rate set to $0.0038$ with batch size $8$ for both architectures. Because the teacher's predictions are unstable at the early stage of the training, we apply the Gaussian ramp-up to the consistency loss weight $\beta$ in~\eqref{eq:main_loss}. 
For both datasets, the supervised loss is the cross-entropy loss.




\textbf{Evaluation metrics.}  Following  previous papers~\cite{chen2021semi, ke2020guided}, we report the mean Intersection-over-Union (mIoU) for validation set 
for both datasets. All the results are based on the single scale inference.

\subsection{Results on Different Partition Protocols}
\label{sec:4.2}
In this section, following~\cite{ke2020guided, chen2021semi}, we evaluate our method based on sub-sampling the datasets with ratio $1/n$ for the labelled set and $(1 - 1/n)$ unlabelled set. Specifically, in the \textit{Pascal VOC 2012} dataset, we split the entire training set (with $10,582$ images) with the ratios of $1/16$, $1/4$, $1/8$, $1/2$ for be labelled set. In the \textit{Cityscapes}, we similarly utilise the ratios $1/8$, $1/4$, $1/2$ for the labelled set. All the partition protocols are from~\cite{chen2021semi}. 

\begin{figure}[t]
     \begin{subfigure}[b]{0.5\textwidth}
         \centering    
         \includegraphics[width=.48\textwidth]{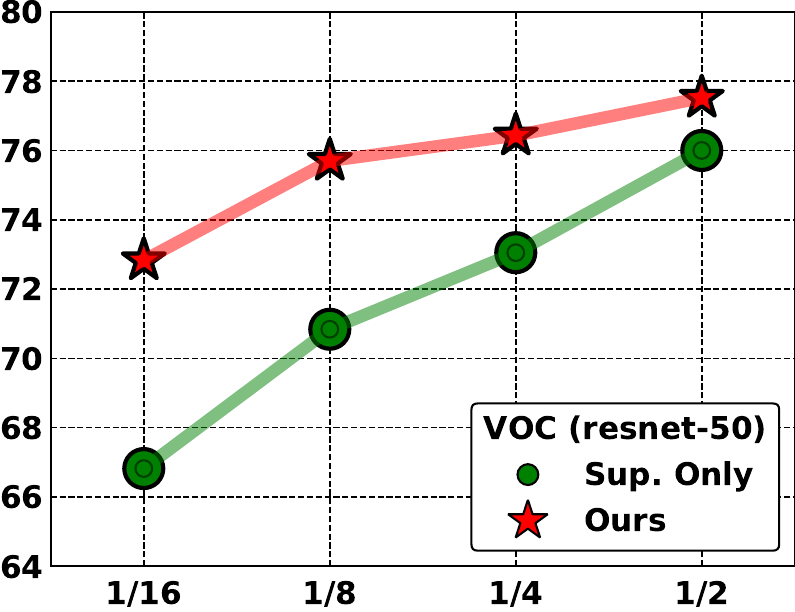}
         \includegraphics[width=.48\textwidth]{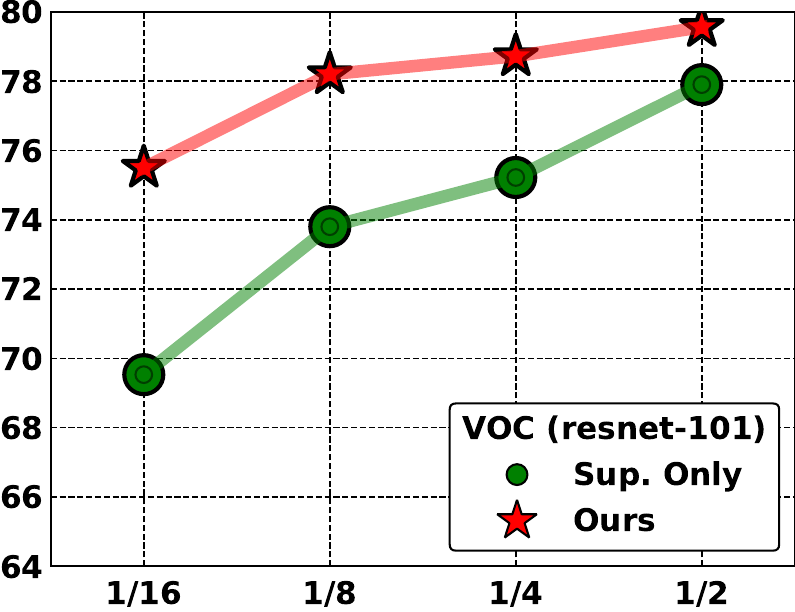}
         \caption{Pascal VOC 2012 dataset}
     \end{subfigure}
     \begin{subfigure}[b]{0.5\textwidth}
         \centering    
         \includegraphics[width=.48\textwidth]{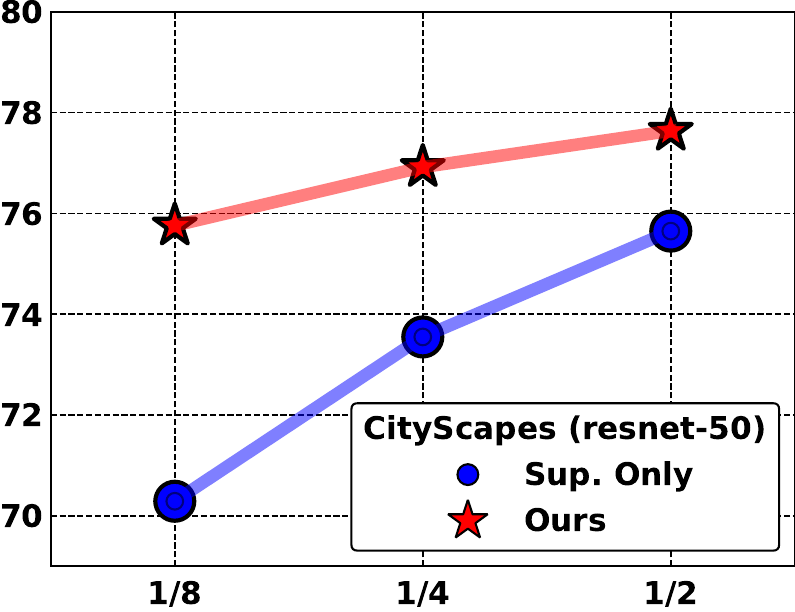}
        \includegraphics[width=.48\textwidth]{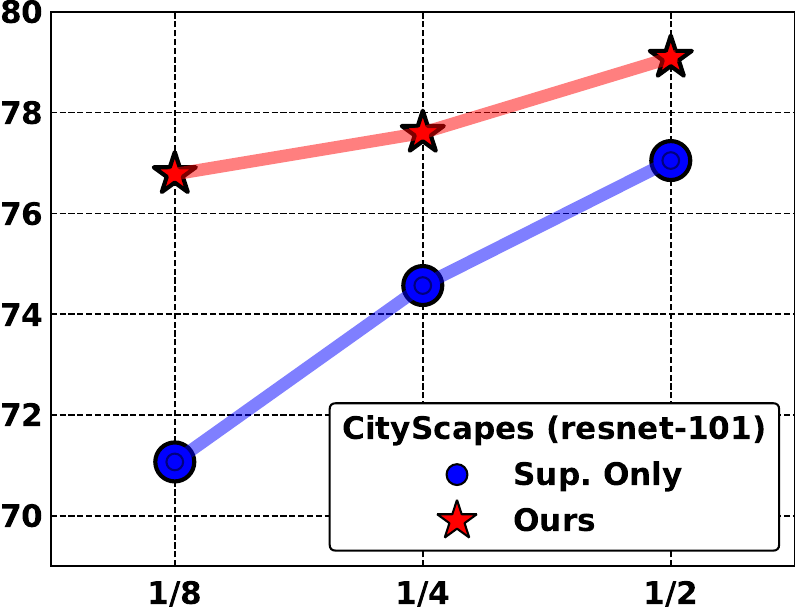}
         \caption{Cityscapes dataset}
     \end{subfigure}
    \caption{\textbf{Improvements over the supervised baseline.} mIoU vs. partition protocols results of our approach and the supervised baseline on  \textit{Pascal VOC 2012} (top) and \textit{Cityscapes} (bottom) using ResNet50 and ResNet101 backbones and DeepLabV3+. 
    } 
    \label{fig:supervised_baseline}

\vspace{-5pt}
\end{figure}

\textbf{Improvements over Supervised Baselines.} We first compare our results with fully supervised learning (trained with the same ratio of labelled set) using DeepLabV3+ architecture, and illustrate the improvements in  Fig.~\ref{fig:supervised_baseline}. This figure demonstrates that our approach successfully exploits unlabelled data, with a dramatic performance boost. On \textit{Pascal VOC 2012}, Fig.~\ref{fig:supervised_baseline}-(a) shows that our approach outperforms the supervised baseline by a large gap, especially for small labelled partitions. Specifically, in the $1/16$ ratio (with $662$ labelled images), our approach surpasses the fully supervised baseline with $6.01\%$ and $5.97\%$ for the ResNet50 and ResNet101, respectively. On the other settings, our approach also shows consistent improvements 
between $2\%$ and $5\%$ for ratios $1/8$, $1/4$, and $1/2$.
On \textit{Cityscapes}, we use the \textbf{sliding evaluation}
to evaluate our final results following~\cite{chen2021semi}. Fig.~\ref{fig:supervised_baseline}-(b) shows that our approach surpasses the supervised baseline by $2\%$ and $6\%$ 
for ResNet50 and ResNet101 for all protocols. 

\textbf{Comparison to SOTA.} For \textit{Pascal VOC 2012}, Tab.~\ref{table:partition-voc} shows that our approach is the best for all partition protocols, using  DeepLabV3+ and ResNet50 and ResNet101 backbones. Comparing to French et al.~\cite{french2019semi}, which is considered a strong baseline, our approach improves by $3\%$ to $6\%$ in all cases. Our approach also provides a significant boost for the original MT in all cases.
The results also show that our approach is better than the current SOTA CPS~\cite{chen2021semi} by around $1\%$ to $2\%$ for all cases. In some partition protocols, our approach is  better than the CPS~\cite{chen2021semi} with fewer labelled samples. For example, our approach trained with $1,323$ labelled images outperforms CPS~\cite{chen2021semi} trained with $2,646$ labelled images using both backbones. This demonstrates that our perturbed and strict mean teachers yield more accurate results than any other method in the field.
On \textit{Cityscapes}, we use the settings from~\cite{lai2021semi} and show results that use similar
settings (in terms of image resolution, batch size, and supervised loss function) for fair comparison.
Our approach outperforms CAC~\cite{lai2021semi} by nearly $4.6\%$ and $2.4\%$ for the $1/8$ and $1/4$ partition protocols. The sliding evaluation also boosts our performance by approximately $1.30\%$ for all the ratios. This shows that the the sliding process improves the performance of our approach in large resolution images. 
\subsection{Results on Official Labelled Set of Pascal VOC 2012}
\label{sec:4.3}
In this section, we report the results based on the official $1,464$ high quality labelled data of the \textit{Pascal VOC 2012}. 
We compare our approach for both PSPNet and DeepLabV3+ architectures. 
Table~\ref{label:official_voc} shows that our approach outperforms the SOTA methods for each architecture and backbone settings. For PSPNet, our approach outperforms DARS~\cite{he2021re} by $0.7\%$ mIoU and CCT~\cite{ouali2020semi} by $5.19\%$. In the  experiments, our approach outperforms other SOTA approaches by a large gap. Yuan et. al~\cite{yuan2021simple} only utilize the single network to produce the pseudo labels in a self-training manner. For instance, our approach outperforms Yuan et. al~\cite{yuan2021simple} by $5.01\%$ mIoU, which shows the value of our teachers and student network and several perturbation strategies, compared with their self-training single network approach. 

\textbf{Few-supervision study.} We subsample with partitions $1/2$, $1/4$, $1/8$ and $1/16$ using the official $1,464$ labelled images. The remaining data are combined with the augmented set~\cite{hariharan2011semantic} (around $9K$ images) to be the unlabelled data in the experiments. 
Table~\ref{table:few_supervision} shows that our approach yields the best mIoU results for all cases. For example, our approach outperforms CPS~\cite{chen2021semi} by $4.86\%$ for $366$ labelled images. 
We argue that our effective perturbations allowed the better generalisation of our model under this limited labelled data conditions.

\begin{table}
\centering
\caption{\textbf{Comparison using  the official ($1,464$) labelled images} on the Pascal VOC 2012 under different network settings. Best results are in bold.}
\resizebox{0.92\linewidth}{!}{
\begin{tabular}{!{\vrule width 1pt}l!{\vrule width 1pt}ll!{\vrule width 1pt}l!{\vrule width 1pt}} 
\specialrule{1pt}{0pt}{0pt}
Method     & Architecture & Backbone   & mIoU  \\ 
\specialrule{1pt}{0pt}{0pt}
CCT~\cite{ouali2020semi}
& PSPNet       & ResNet50  &  69.40     \\
DARS~\cite{he2021re}
& PSPNet       & ResNet50  &  73.89     \\
ours       & PSPNet       & ResNet50  &     \textbf{74.59} \\ 
\specialrule{1pt}{0pt}{0pt}
CAC~\cite{lai2021semi}   & DeeplabV3+ & ResNet50   &  74.50     \\
ours       & DeeplabV3+   & ResNet50  &       \textbf{78.08} \\
\specialrule{1pt}{0pt}{0pt}
PseudoSeg~\cite{zou2020pseudoseg}
& DeeplabV3+   & ResNet101  &  73.20    \\
Yuan et al.~\cite{yuan2021simple}
&DeeplabV3+ & ResNet101   &  75.00     \\
ours       & DeeplabV3+   & ResNet101 &    \textbf{80.01}   \\
\specialrule{1pt}{0pt}{0pt}
\end{tabular}
} \label{label:official_voc}
\end{table}

\begin{table}[t!]
\caption{\textbf{Comparison with SOTA approaches with few-supervision using  the official ($1,464$) labelled images} on the Pascal VOC 2012. Our approach follows the same protocols as CPS~\cite{chen2021semi} and PseudoSeg~\cite{zou2020pseudoseg}. Best results are in bold.}
\centering
\resizebox{.9\linewidth}{!}{
\begin{tabular}{!{\vrule width 1pt}c!{\vrule width 1pt}c!{\vrule width 1pt}c!{\vrule width 1pt}c!{\vrule width 1pt}c!{\vrule width 1pt}} 
\specialrule{1pt}{0pt}{0pt}
Method     & 732    & 366   & 183  & 92     \\ 
\specialrule{1pt}{0pt}{0pt}
AdvSemSeg~\cite{hung2018adversarial}  & 65.27 & 59.97 & 47.58 & 39.69  \\
CCT~\cite{ouali2020semi}        & 62.10 & 58.80 & 47.60 & 33.10  \\
MT~\cite{tarvainen2017mean}         & 69.16 & 63.01 & 55.81 & 48.70  \\
GCT~\cite{ke2020guided}        & 70.67 & 64.71 & 54.98 & 46.04  \\
VAT        & 63.34 & 56.88 & 49.35 & 36.92  \\
French et al.~\cite{french2019semi} & 69.84 & 68.36 & 63.20 & 55.58  \\
PseudoSeg~\cite{zou2020pseudoseg}  & 72.41 & 69.14 & 65.50 & 57.60  \\
CPS~\cite{chen2021semi}        & 75.88 & 71.71 & 67.42 & 64.07  \\ 
\specialrule{1pt}{0pt}{0pt}
ours       &  \textbf{78.42} &  \textbf{76.57} & \textbf{69.58} & \textbf{65.80}       \\
\specialrule{1pt}{0pt}{0pt}
\end{tabular}
}
\label{table:few_supervision}
\end{table}

\subsection{Ablation Study}
\label{sec:4.4}

In this section, we study the roles of the confidence weighted CE loss (conf-CE), T-VAT perturbation and auxiliary teacher (AT) of our approach. All the experiments are run on  \textit{Pascal VOC 2012} under $1/8$ ratio, and we use DeeplabV3+ to evaluate our results.  
\begin{table}[h]
\centering
\caption{\textbf{Ablation study} using the $1/8$ labelled ratio on \textit{Pascal VOC 2012} under DeeplabV3+ architecture.}
\resizebox{0.92\linewidth}{!}{
\begin{tabular}{!{\vrule width 1pt}c|c|c|c!{\vrule width 1pt}c|c!{\vrule width 1pt}c} 
\specialrule{1pt}{0pt}{0pt}
\multirow{2}{*}{MT} & \multirow{2}{*}{conf-CE} & \multirow{2}{*}{T-VAT} & \multirow{2}{*}{AT} & \multicolumn{2}{c!{\vrule width 1pt}}{Backbone}  \\ 
\cline{5-6}
                               &                                      &                                               &                                                   &  ResNet-50 & ResNet-101          \\ 
\hline
                              \checkmark  &                                     &                                               &                                                   &        71.49  & 73.50                   \\
                              \checkmark  &                                     \checkmark  &                                               &                                                   &          73.79 & 76.39                    \\
                               \checkmark &                                     \checkmark  &                                             \checkmark   &                                                   &        74.87  &    77.36                \\
                            
                            
                            \checkmark   &                                    \checkmark   &                                             \checkmark   &                                                 \checkmark   &        \textbf{75.70}  &    \textbf{78.20}                \\
\specialrule{1pt}{0pt}{0pt}
\end{tabular}}
\label{table:ablation}
\vspace{-15pt}
\end{table}
\begin{figure}[ht!]
    \centering     
    \begin{subfigure}[b]{0.49\linewidth}
         \centering    
            \includegraphics[width=1.\linewidth]{ 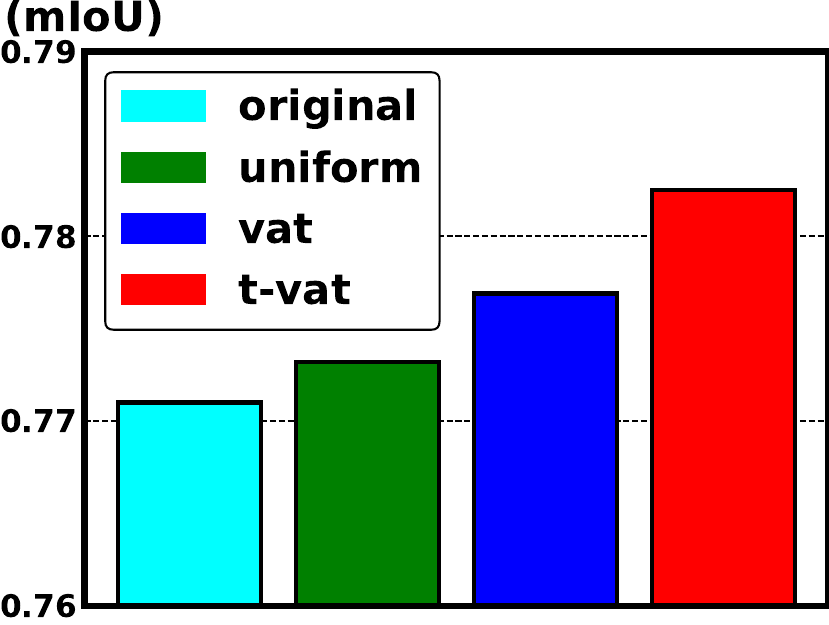}
        \caption{}
    \end{subfigure}
    \begin{subfigure}[b]{0.490\linewidth}
        \centering    
        \includegraphics[width=1.\linewidth]{ 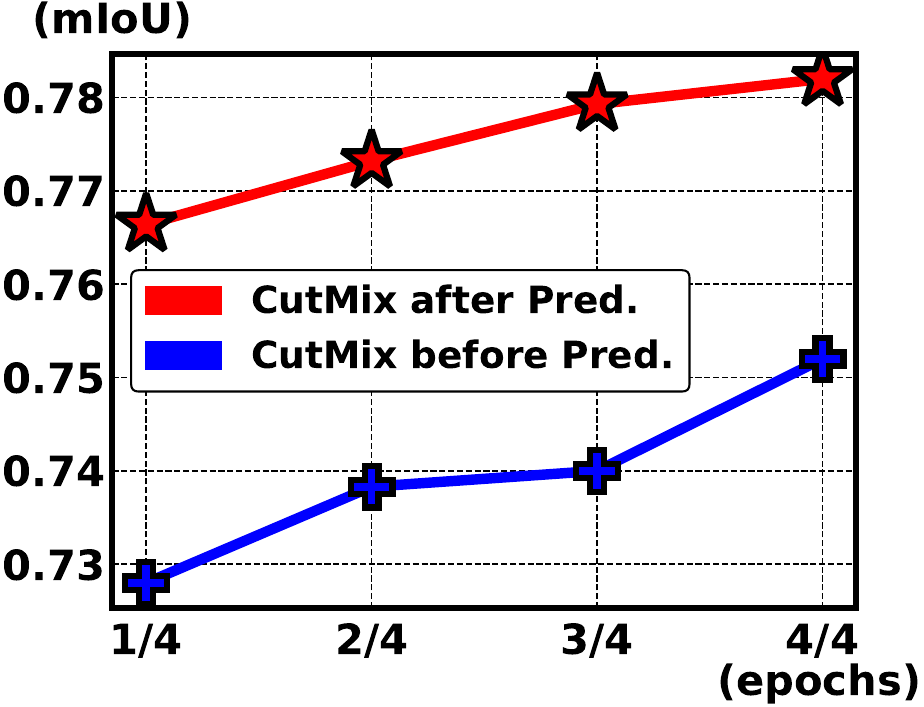}
        \caption{}
     \end{subfigure}
    \caption{\textbf{T-VAT and CutMix Perturbation Effectiveness.} (a) mIoU under different feature perturbations. (b) mIoU of validation set during training for CutMix applied before or after prediction, as described in~\eqref{eq:cutmix_loss}. 
    } \vspace{-5pt}
    \label{fig: perturbation}
\end{figure}
Table~\ref{table:ablation} demonstrates the improvements of each component mentioned above, where we use MT~\cite{tarvainen2017mean} trained with the input image perturbations from Sec.~\ref{sec:consistency_loss} and MSE loss as baseline. 
We note that by replacing MSE by our conf-CE increases mIoU by $2.30\%$ and $2.81\%$ for the ResNet50 and ResNet101. T-VAT perturbation yields nearly $1\%$ improvements, showing the effectiveness of our proposed feature perturbation. The more accurate predictions by the auxiliary teacher allows a further improvement of $0.83\%$ and $0.84\%$ for the two backbones.


\textbf{T-VAT perturbation.} Fig.~\ref{fig: perturbation}-(a) shows the performance under different types of feature perturbations, namely: original (no feature perturbation), uniform (feature noise randomly sampled from uniform distribution), vat (VAT noise learned from the student model), t-vat (T-VAT noise learned from the teacher model as in~\eqref{eq:teacher_ensemble_adversarial_result}). Our proposed T-VAT outperforms uniform and VAT perturbations by $0.93\%$ and $0.62\%$, respectively. Additionally, it also surpasses original by $1.10\%$. 

\textbf{Empirical results of the CutMix before or after prediction, as described in~\eqref{eq:cutmix_loss}.} In Fig~\ref{fig: perturbation}-(b), we show the mIoU results on the validation set during training epochs. Applying CutMix before predictions may introduce extra semantic complexity and yield inaccurate pseudo-labels, leading to ineffective optimisation. In contrast, the result indicates that applying the CutMix after the prediction improves mIoU by around $3\%$. 


\begin{table}[h!]
\centering 
\caption{\textbf{Comparison using  the official ($1,464$) labelled images with combined pseudo-label and consistency-based losses} on the Pascal VOC 2012 under different network settings. Best results per architecture are in bold. 
}\resizebox{.95\hsize}{!}{$%
\begin{tabular}{lll!{\vrule width 1pt}l} 

\specialrule{1pt}{0pt}{0pt}
Method  & Architecture & Backbone   & mIoU  \\ 
\specialrule{1pt}{0pt}{0pt}
CCT~\cite{ouali2020semi} &       PSPNet       &      ResNet50      &   73.2    \\
ours    & PSPNet         & ResNet50  & \textbf{75.74}       \\
\specialrule{1pt}{0pt}{0pt}
AdvCAM~\cite{lee2021anti} &  Deeplabv2            &    ResNet101        & 77.8      \\
PseudoSeg~\cite{zou2020pseudoseg} & DeeplabV3+             &    ResNet50        &   73.8    \\ 
CAC~\cite{lai2021semi}  & DeeplabV3+             &    ResNet50       &  76.1    \\ 
ours    & DeeplabV3+     & ResNet50  &       78.71 \\
ours    & DeeplabV3+     & ResNet101 &       \textbf{81.19} \\
\specialrule{1pt}{0pt}{0pt}
\end{tabular} 
$}
\label{table:cam_exp}
\end{table}

\begin{figure}[h!]
    \centering
    \includegraphics[width=.95\linewidth]{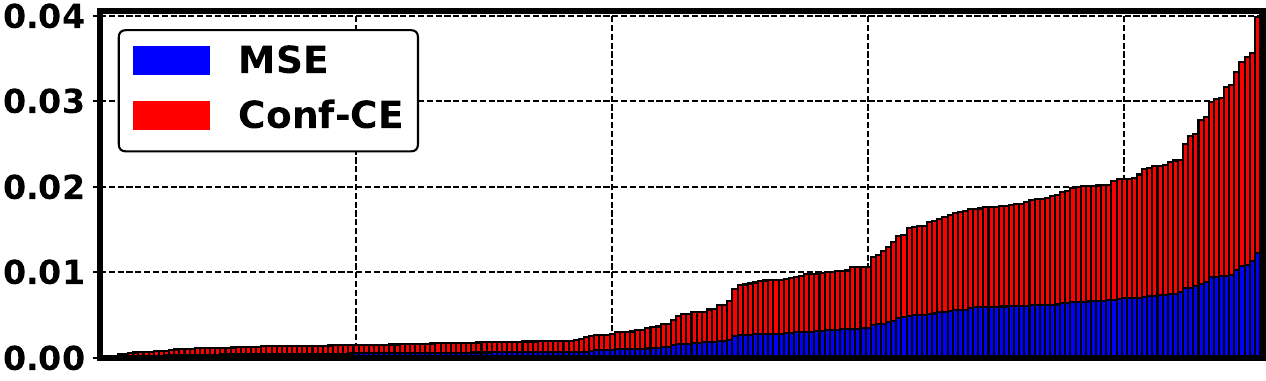}
    \caption{\textbf{Different gradient magnitudes}. Average gradient magnitudes per layer of the student model after being trained with \textbf{MSE} and \textbf{Conf-CE} losses, at the last stages of training. 
    }\vspace{-8pt}
    \label{fig:gradient_diff}
\end{figure}
\textbf{Average gradient magnitudes difference between MSE and Conf-CE.} Fig.~\ref{fig:gradient_diff} shows the average gradient magnitudes per layer of the student model after being trained with MSE and Conf-CE losses to optimise consistency in~\eqref{eq:loss_semi}, at the last stages of training ($70^{th}$ out of 80 epochs). Note that our Conf-CE shows larger gradient magnitudes than MSE~\cite{ke2020guided, chen2021semi, french2019semi}, suggesting that it can lead to stronger convergence than MSE. 

\subsection{Qualitative Results}

\begin{figure}
    \centering
     \begin{subfigure}[b]{0.19\linewidth}
        \includegraphics[width=\linewidth]{ 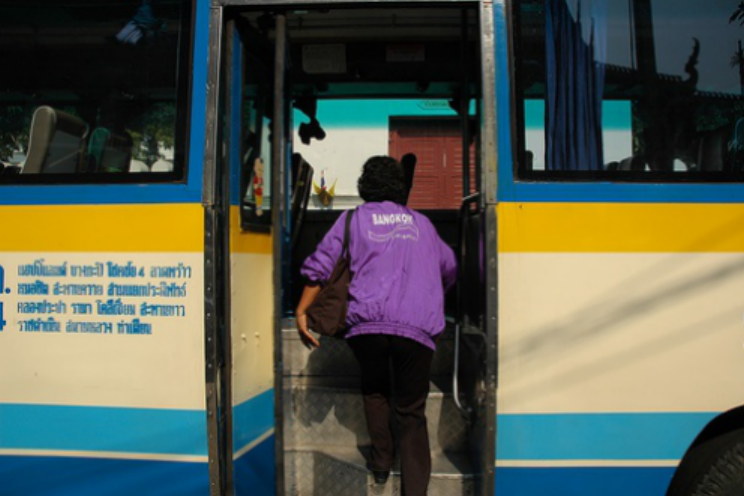}
        \includegraphics[width=\linewidth]{ 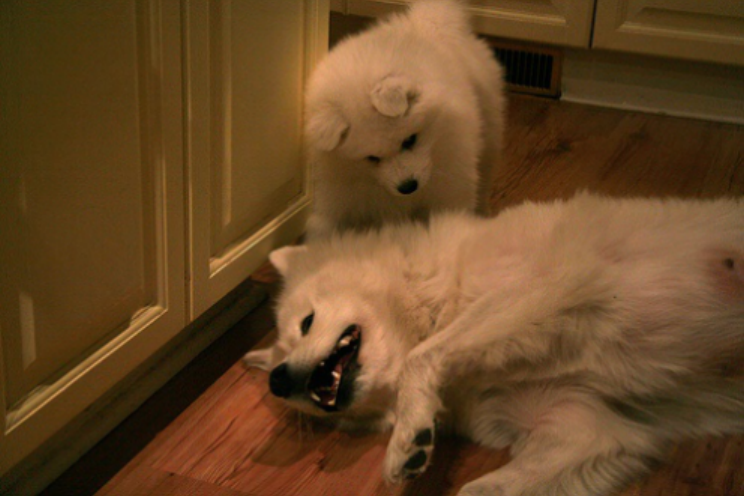}
        \includegraphics[width=\linewidth, height=62pt]{ 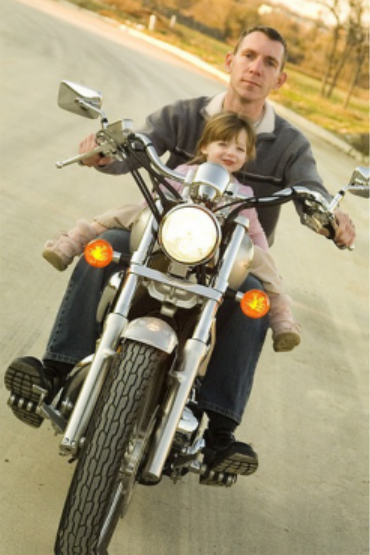}
        \caption{}
     \end{subfigure}
    \begin{subfigure}[b]{0.19\linewidth}
        \includegraphics[width=\linewidth]{ 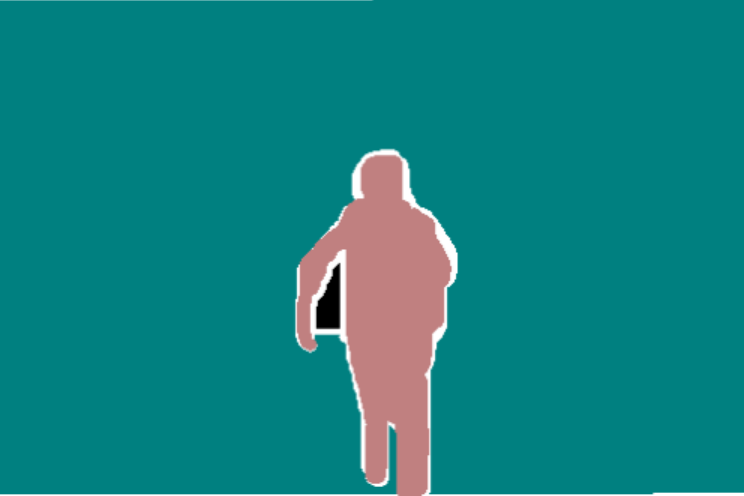}
        \includegraphics[width=\linewidth]{ 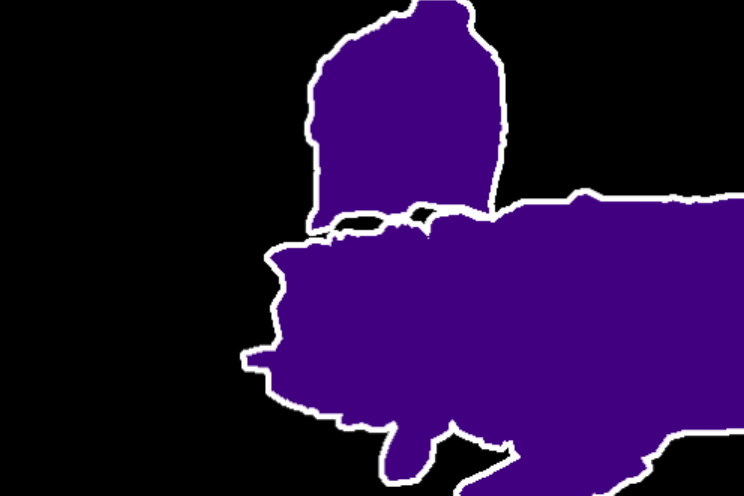}
        \includegraphics[width=\linewidth, height=62pt]{ 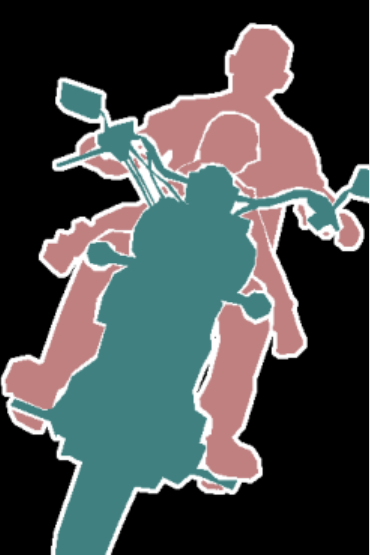}
        \caption{}
    \end{subfigure}
     \begin{subfigure}[b]{0.19\linewidth}
        \includegraphics[width=\linewidth]{ 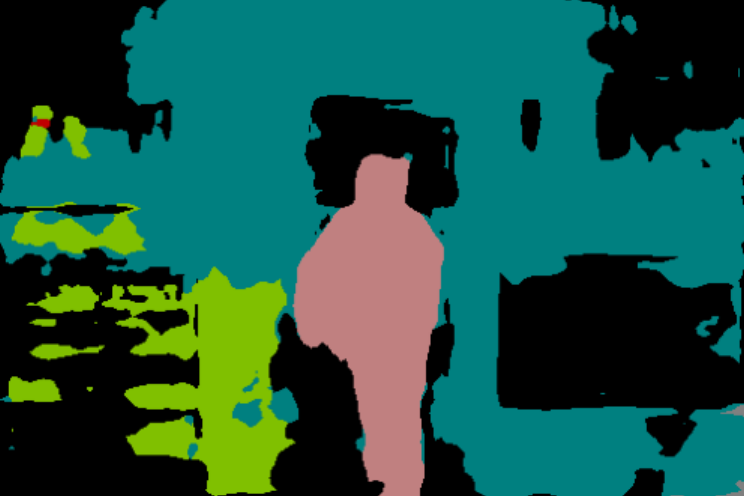}
        \includegraphics[width=\linewidth]{ 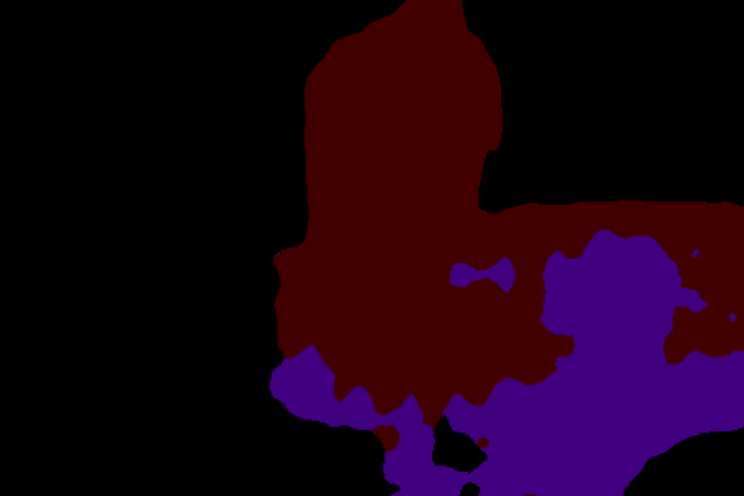}
        \includegraphics[width=\linewidth, height=62pt]{ 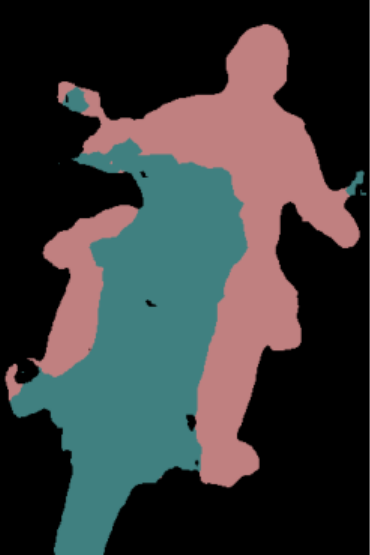}
    \caption{}
    \end{subfigure}
     \begin{subfigure}[b]{0.19\linewidth}
        \includegraphics[width=\linewidth]{ 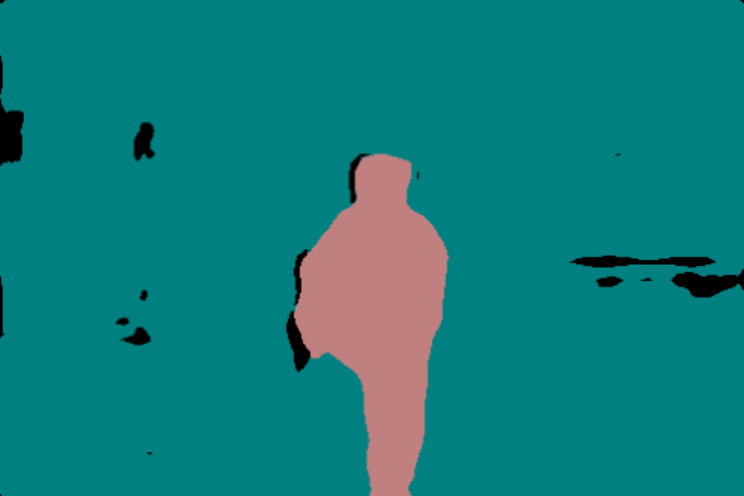}
        \includegraphics[width=\linewidth]{ 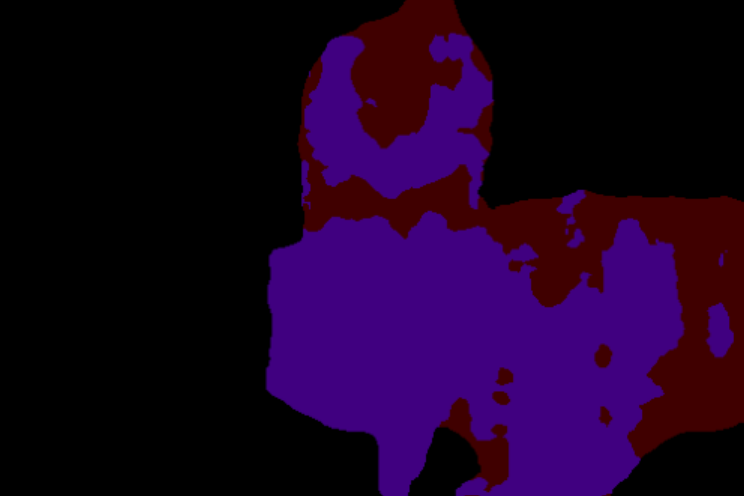}
        \includegraphics[width=\linewidth, height=62pt]{ 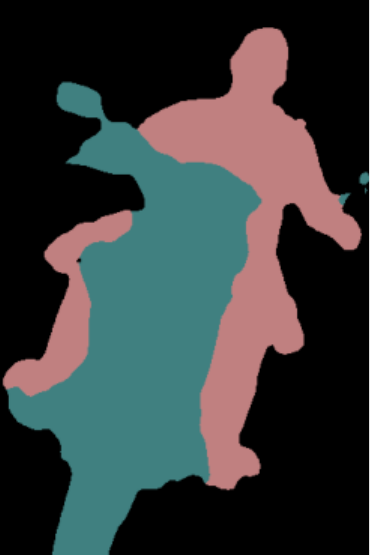}
    \caption{}
    \end{subfigure}
    \begin{subfigure}[b]{0.19\linewidth}
        \includegraphics[width=\linewidth]{ 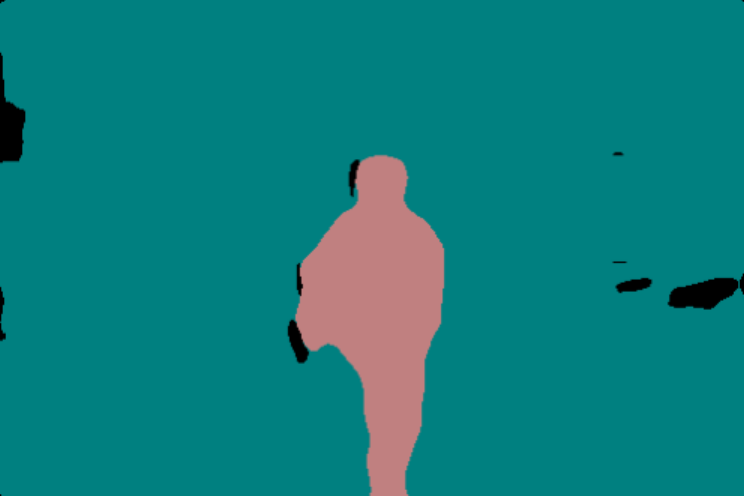}
        \includegraphics[width=\linewidth]{ 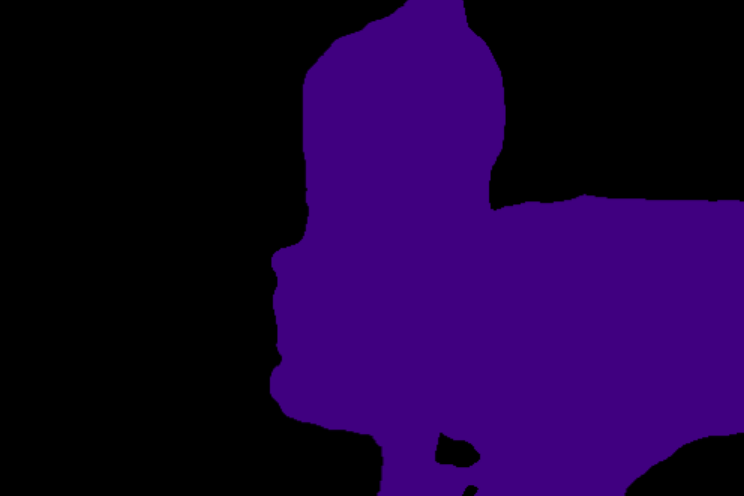}
        \includegraphics[width=\linewidth, height=62pt]{ 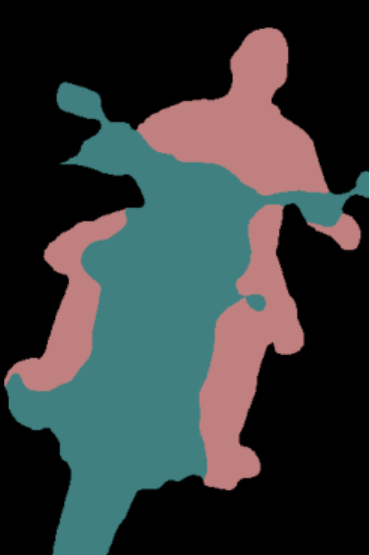}
    \caption{}
    \end{subfigure}
    \caption{\textbf{Qualitative results} from \textit{Pascal VOC 2012}. (a) input images, (b) ground truth, (c) supervised baseline results, (d) student results and (e) results by our approach. 
    } \vspace{-10pt}
    \label{fig:quality_results}
\end{figure}

Figure~\ref{fig:quality_results} shows the supervised, student and mean teachers results on \textit{Pascal VOC 2012} images. The supervised results display the worst accuracy in column (c), caused by the limited labelled training samples. Our final results in column (e) significantly improves the baseline performance, which demonstrates the effectiveness of our approach. Moreover, our final results in column(e) are also more accurate than the student results in column (d).



\subsection{Combining Pseudo-label and Consistency-based losses}
\label{sec:pseudo_label_consistency}

Current consistency-based methods~\cite{ouali2020semi,huang2018weakly,lee2019ficklenet,lai2021semi} also include a pseudo-labelling loss, involving the use of class activation maps (CAM) from the model to generate pseudo-labels $\bar{\mathbf{y}}$. We follow a similar strategy as in CCT~\cite{ouali2020semi} and add the CAM loss below to the cost function in~\eqref{eq:main_loss} to 
train the student model:
\begin{equation} 
\begin{aligned}
\ell_{cam}(&\mathcal{D}_U,\theta^s) = \\ &\frac{1}{|\mathcal{D}_U||\Omega|}\sum_{\mathbf{x}\in\mathcal{D}_U}
\sum_{\omega \in \Omega} (1-c(\omega))\ell(\bar{\mathbf{y}}(\omega),
f_{\theta^{s}}(\mathbf{x})(\omega)),
    \label{eq:loss_cam}
\end{aligned}
\end{equation}
where $c(\omega)$ is the segmentation confidence from the teachers defined in~\eqref{eq:loss_semi}, and $\ell(.)$ is the CE loss. 
Hence, the non-confident predictions from the teachers are then handled by this CAM loss in~\eqref{eq:loss_cam}.

We run experiments based on the official $1,464$ labelled images on \textit{Pascal VOC 2012} with the additional  $\approx 9K$ images used to minimise the CAM loss in~\eqref{eq:loss_cam}. 
Results on Tab.~\ref{table:cam_exp} show that our method outperforms all  previous works that use a similar strategy~\cite{ouali2020semi, lai2021semi, zou2020pseudoseg}. 
Moreover, the addition of this CAM loss in~\eqref{eq:loss_cam} to our cost function boosts the performance by $0.63\%$ and $1.18\%$ for the two backbones under DeeplabV3+ architecture, compared with our previous results in Tab.~\ref{label:official_voc}. 

\vspace{-3pt}
\section{Conclusion}

In this paper, we proposed a new consistency-based semi-supervised semantic segmentation method. Among our contributions, we introduced a new MT model, based on multiple mean teachers and a student network, which shows more accurate predictions for unlabelled images that facilitate consistency learning, allowing us to use a stricter confidence-based CE loss than the original MT's MSE loss.
This more accurate predictions also allowed us to use a challenging combination of network, feature and input image perturbations that showed better generalisation.
Furthermore, we proposed a new adversarial feature perturbation, called T-VAT, that further improved the generalisation of our approach. 
Our method outperforms previous methods on Pascal VOC 2012 and Cityscapes, becoming the new SOTA for the semi-supervised semantic segmentation field. 
Regarding the limitations of our model, it can be argued that the strict Conf-CE loss has the potential to overfit the remaining prediction mistakes, so we will focus on improving the robustness of the Conf-CE loss.
Another limitation that we plan to address is to work on an approach that can handle  high-resolution images without using the time-consuming sliding evaluation.

{\small
\bibliographystyle{ieee_fullname}
\bibliography{egbib}
}

\end{document}